\documentclass[sigconf]{acmart}

\usepackage[utf8]{inputenc}
\usepackage{subfig}
\usepackage{multirow}
\usepackage{chngcntr}

\def\BibTeX{{\rm B\kern-.05em{\sc i\kern-.025em b}\kern-.08emT\kern-.1667em\lower.7ex\hbox{E}\kern-.125emX}}
    
\acmConference[KDD 2020 Workshop]{AI for fashion
The fifth international workshop on fashion and KDD}{August 2020}{San Diego, California-USA}
\setcopyright{none}

\acmPrice{15.00}
\acmDOI{}
\acmISBN{} 
\begin{document}

%
\title{Product age based demand forecast model for fashion retail}

%

\author{Rajesh Kumar Vashishtha}
\affiliation{\institution{TCS Research}
	\city{Chennai}
	\state{India}
	\postcode{123456}
}
\email{r.vashishtha@tcs.com}

\author{Vibhati Burman}
\affiliation{\institution{TCS Research}
	\city{Chennai}
	\state{India}
	\postcode{123456}
}
\email{vibhati.b@tcs.com}

\author{Rajan Kumar}
\affiliation{\institution{TCS Research}
	\city{Chennai}
	\state{India}
	\postcode{123456}
}
\email{rajan.7@tcs.com}

\author{Srividhya Sethuraman}
\affiliation{\institution{TCS Research}
	\city{Chennai}
	\state{India}
	\postcode{123456}
}
\email{srividhya.sethuraman1@tcs.com}

\author{Abhinaya R Sekar}
\affiliation{
	\institution{TCS Research}
	\city{Chennai}
	\state{India}
	\postcode{123456}
}
\email{abhinayar.sekar@tcs.com}

\author{Sharadha Ramanan}
\affiliation{
	\institution{TCS Research}
	\city{Chennai}
	\state{India}
	\postcode{123456}
}
\email{sharadha.ramanan@tcs.com}

%
\renewcommand{\shortauthors}{Rajesh, Vibhati, Rajan, Srividhya, Abhinaya and Sharadha}

%
\begin{abstract}
Fashion retailers require accurate demand forecasts for the next season, almost a year in advance, for demand management and supply chain planning purposes. Accurate forecasts are important to ensure retailers' profitability and to reduce environmental damage caused by disposal of unsold inventory. It is challenging because most products are new in a season and have short life cycles, huge sales variations and long lead-times. In this paper, we present a novel product age based forecast model,  where product age refers to the number of weeks since its launch, and show that it outperforms existing models. We demonstrate the robust performance of the approach through real world use case of a multinational fashion retailer having over 300 stores, 35k items and around 40 categories. The main contributions of this work include unique and significant feature engineering for product attribute values, accurate demand forecast 6-12 months in advance and extending our approach to recommend product launch time for the next season. We use our fashion assortment optimization model to produce list and quantity of items to be listed in a store for the next season that maximizes total revenue and satisfies business constraints. We found a revenue uplift of 41\% from our framework in comparison to the retailer's plan. We also compare our forecast results with the current methods and show that it outperforms existing models. Our framework leads to better ordering, inventory planning, assortment planning and overall increase in profit for the retailer's supply chain.
\end{abstract}

%
%

%
\keywords{Demand forecast, Fashion assortment, Assortment optimization, Supply chain planning, Age based algorithm, Fashion industry, New item forecast, Item launch time.}

\maketitle

\section{Introduction}
Fashion is a \$1.7 trillion value industry. It is one of the fastest growing sectors and also one of the largest waste producers globally due to over production and returns. Fashion retailers face the challenge of picking the right mix from the 150 billion garments being produced every year. Study shows that 30\% of these are never sold resulting in a loss of \$210 billion per year globally due to inventory distortion \cite{WinNT}. Overstocking leads to disposing of over 12.8 million tons of clothing to landfills annually. Thus, accurate demand forecasting of fashion items is very crucial for fashion retailers.

With rapidly changing consumer preferences, macro and micro influences, short life cycles, long lead times and 'see now buy now'  trends, it has become increasingly difficult to predict demand accurately. Fashion items have a wide variety of attributes and mostly new items to be sold in a season. Moreover, real world data is sparse, noisy and incomplete which adds to the challenge of forecasting. Furthermore, retailers need to forecast sales for fashion items in a store for the next season at the end of the current season itself, i.e., 6-12 months in advance.

Current statistical and other forecast models predict well only when there is historical data, few fixed attributes, regular sales patterns and longer lifecycles. Therefore, current models do not predict well for fashion items.

In this paper, we present a novel age-based prediction model that accurately forecasts demand for fashion items for the next season. We have modeled sales of a fashion item in a store as a function of its age, attributes, selling price, temporal and store features. Novel feature engineering is also introduced in this work.

In addition, it is valuable to leverage accurate forecasts and integrate it with a larger framework. They have to be integrated into a larger framework in supply chain planning consisting of key decisions including ordering from vendors, inventory planning, season planning, assortment planning and determining new item launch time. Thus, we need a forecast method that gives not just low forecast errors but also can be malleable to be fit inside a larger framework of a retailers' sales perspective and supply chain.

The bigger problem for the fashion retailers is not only to forecast but also to know when they need to launch the new items in the upcoming season. This is another problem that we solve in this paper. Finally, assortment planning, based on the forecasts, produces list of items and number of units to be listed in a store for the next season that maximizes total revenue and satisfies constraints based on retailer's strategic rules.

The organization of this paper is given as follows: Section 2 presents a review of related work. In Section 3, characteristics of fashion retailer's data is summarized. Section 4 discusses the methodology followed. Section 5 presents experiments done and explains the results obtained. Finally, in Section 6 we discuss the conclusions of the work and suggest the direction of future research work.

\section{Related Work }
Traditional statistical forecasting techniques such as exponential smoothing, ARIMA and regression models do not give accurate fashion forecasts due to irregular patterns and high variability of sales for fashion items. Time series models are based on forecasts obtained from previous year's sales of similar items. Fashion items are launched in different months within a season. Hence, for a new item, it becomes increasingly difficult to identify the similar items from sales history as those similar items might have been launched across different months. Therefore, time series models are not suitable in this scenario. Models involving Bayesian methods, Machine Learning (ML) and Neural Networks (NN) perform better than statistical techniques; however their accuracy is still not satisfactory in case of items with short life cycles and no historical sales  \cite{gelman2013bayesian,liu2013sales}. 

Existing Artificial Intelligence (AI) techniques such as Artificial Neural Networks (ANN) when applied to our real world use case, do not give promising results. This is due to the launch of new fashion items across different months within a season and also since items have a rich attribute set wherein, about 20-30\% of attribute values are new at every launch. Other models involving ANN architectures such as the ones implementing Fuzzy Logic \cite{mastorocostas2001constrained} and Evolutionary NN \cite{au2008fashion} require substantial amount of time for training. In the case of extreme learning machines, instability arising due to variable outputs on each run is a huge impediment for its deployment in industry.

In \cite{nenni2013demand}, the authors highlight the challenges in forecasting demand for fashion items that are subject to short selling seasons, high level of uncertainty and lack of historical data. They point out that color is an important attribute. We have used novel feature engineering on the color attribute in our model (See Section 4.1). In \cite{liu2013sales} the authors discuss hybrid models that combine statistical and AI techniques to provide lot more advantages. Our model considers the average model as the baseline and forecasts of our age based models are compared against the baseline. The authors also highlight that special events or calendar factors are important features but often not considered. We have included these features in our model (see Section 4.1).

In \cite{kharfan2018forecasting}, the authors present an approach comprising of two independent models to handle new fashion items for a U.S based footwear retailer. The first model used regression trees, random forests, K-nearest neighbours and neural networks whereas the second model involved clustering, classification and prediction. In \cite{loureiro2018exploring}, the authors forecast for real world fashion sales data. They note that Deep learning (DL)  models achieve greater accuracy because of domain inputs. Manual intervention is necessary for both these models. Our model, in contrast, is fully automated.

In \cite{wong2010hybrid}, the authors propose a two step hybrid intelligent model- a forecasting component using an improved harmony search algorithm integrated with extreme learning machine and a heuristic fine-tuning to improve predictions. This method has high time-complexity and is not feasible for real world deployment. In \cite{singh2019fashion}, the authors considered historical sales data of an e-commerce company. The model used sales data of an online store and required initial four weeks of sales in a season for training their models. Our model is capable of predicting demand 6-12 months in advance, independent of in-season sales. Furthermore, we have considered sales data of brick and mortar stores and we have employed store clustering. Hence, our approach is different from theirs. 

In \cite{thomassey2006hybrid}, the authors propose an approach of clustering of similar SKUs based on their sales profile and a decision tree classifier to determine the appropriate cluster for a new fashion item. They had tested this on a very small data set of 285 real sales items. Our scenario consists of large-scale data. In \cite{beyer2005profile}, the authors describe a model that has a demand predictor that provides total life cycle demand for a fashion item. Our scenario requires model that predicts demand for each week in a season.

Assortment in fashion is choosing the right depth (number of units) and breadth (variety) of fashion items that maximizes revenue. In \cite{rajaram2001assortment}, the authors models the assortment problem as a non–linear integer programming problem and combines efficient heuristics to make assortment decisions. Consequently, the authors claim that their approach led to increase in profits when applied for a retailer specializing in Women's apparel.

In \cite{farias2017building}, the authors have addressed the problem of inventory planning by using non-parametric choice modelling and an optimization framework. The assortment optimization problem is formulated with the objective of maximising profits with linear constraints. The constraints mentioned are in reference to demand for different products, store level inventory and budget. The authors have tested their approach for a large U.S fashion retailer and they claim that their approach was able to generate 7\% revenue growth.

\section{Data Characteristics}

The data of a multinational fashion retailer, anonymized and used for this study consists of over 300 brick and mortar stores, 40 categories and around 35k items. Data is structured and contains performance and attribute related information for offline stores. In this paper, we give details of application of our model for two cases - Dresses in Women's wear and Kids wear. In this section, we discuss some of the significant insights obtained from the data for Dresses and Kids wear.

\begin{figure}[!ht] 
	\subfloat[Dresses: Sales pattern with Age]{%
		\includegraphics[width=0.4\textwidth]{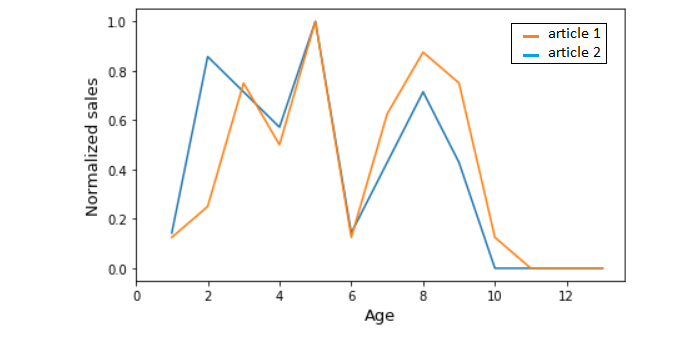} 
		\label{age_dress}
	} 
	\\
	\subfloat[Kids wear: Sales pattern with Age ]{%
		\includegraphics[width=0.4\textwidth]{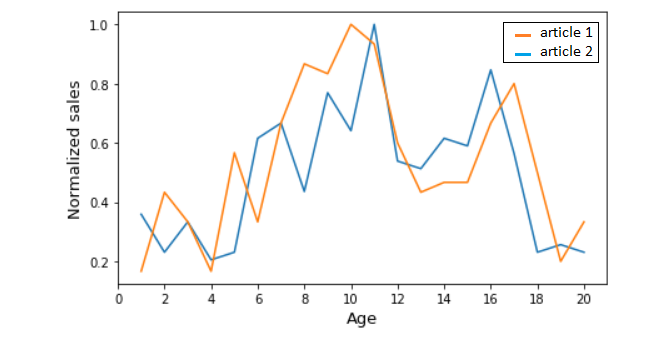}
		\label{age_kids} 
	} 
	\caption{Sales pattern with Age }
	\label{AGE_FIG}
\end{figure}

\begin{figure*}[!ht] 
	
	\subfloat[Dresses: Color]{%
		\includegraphics[height=0.8in,width=0.3\textwidth]{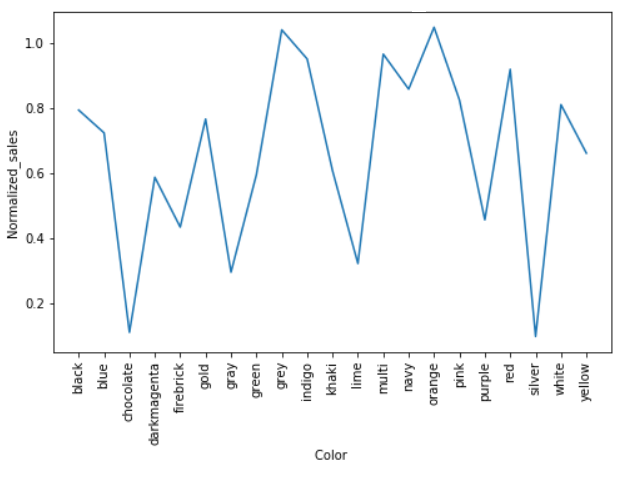} 
		\label{color_dress}
	} 
	\subfloat[Dresses: Sleeve length]{%
		\includegraphics[height=0.8in,width=0.3\textwidth]{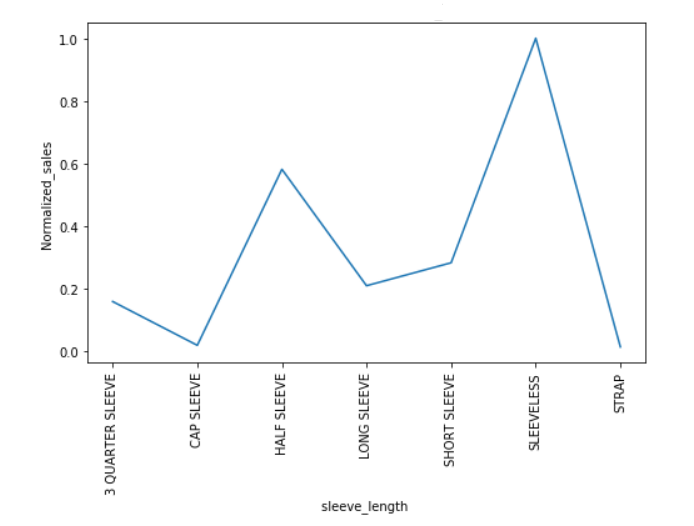} 
		\label{sleeve_dress}
	} 
	\subfloat[Dresses: Selling price]{%
		\includegraphics[height=0.8in,width=0.3\textwidth]{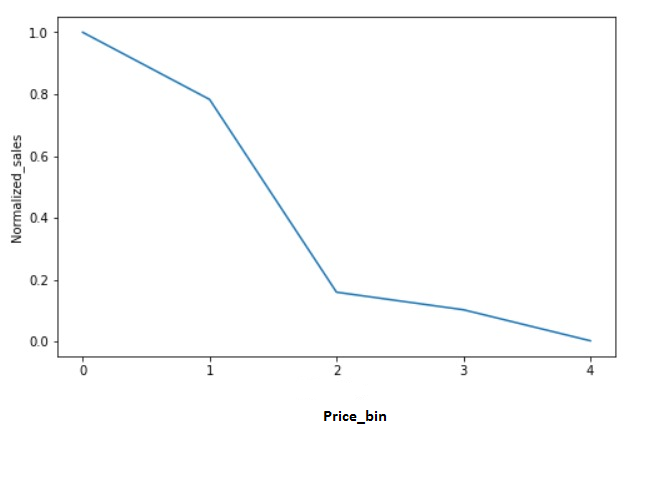} 
		\label{price_dress}
	} 
	\\
	\subfloat[Kids wear: Color]{%
	\includegraphics[height=0.8in,width=0.3\textwidth]{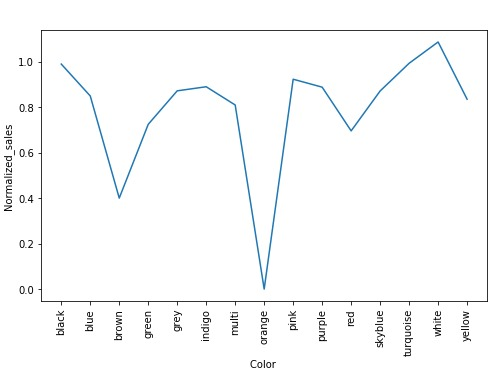} 
	\label{color_kids}
} 
\subfloat[Kids wear: Sleeve length]{%
	\includegraphics[height=0.8in,width=0.3\textwidth]{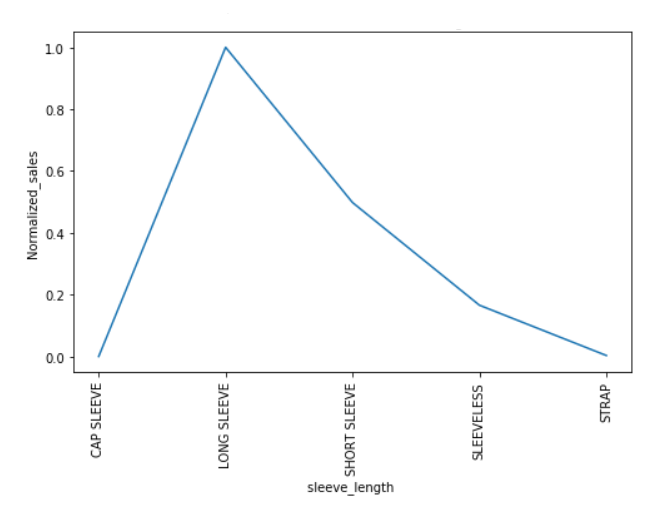} 
	\label{sleeve_kids}
}
\subfloat[Kids wear: Selling price]{%
	\includegraphics[height=0.8in,width=0.3\textwidth]{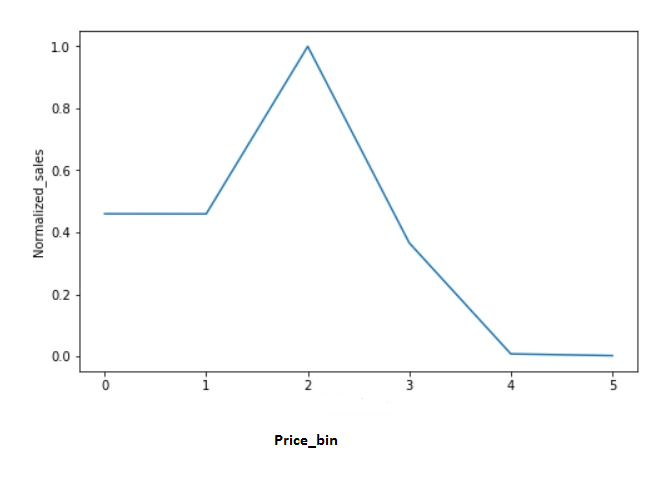} 
	\label{price_kids}
}

\caption{Sales Pattern with Attributes.}
\label{fig:Feature}
\end{figure*}

The product hierarchy for a fashion items is: category, class, sub-class, article and variant. Women's wear and Kids wear are two different categories. Dresses, footwear and trousers are classes within Women's wear. Similarly, tops and pyjamas are classes within Kids wear. Items within Dresses in Women's wear have common attributes such as Sleeve length, Color, Pattern, Fastening type and Neck shape. For Kids wear, items within the category itself share common attributes such as Sleeve length, Color and Fit. Therefore, prediction is done at class level, i.e., for Dresses in Women's wear and at category level for Kids wear. The sub-class describes the specific style pertaining to fashion items. Bodycon, jumpsuit and swing dresses are typical sub-classes for Women's Dresses, and newborn and unisex for Kids wear. An article contains fashion items within a sub-class with more specific style and price related information. The variant level encompasses fashion articles with their respective size variants. Article is also referred as item/product in this work.

Analysis of 3 years of historical data shows that a season consists of almost 6 months. A season is further split into 4 phases. Each phase comprises of 6-7 weeks. Assortment of the articles in each store are updated at the starting of each phase. From the given phase's assortment, only 35-40\% of the articles are carried over to the next phase and many new articles are introduced in the upcoming phase.

In our analysis, we found that around 1000 articles are launched in a season for both Dresses and Kids wear. Also, 85-90\% of these articles are new and hence only 10-15\% of articles have historical sales data. Articles belonging to Dresses and Kids wear have short life cycles. Most of the articles in Dresses have a life cycle of less than 10 weeks while in Kids wear it is less than 15 weeks. Further, the sales of an article is not contiguous as there are many non-sale days in between. For both, most of the articles sell only for 1–2 days in a week at a store but weekly sales are non-zero for the articles within its life cycle. At a variant level, there are many zero sales even at the weekly level. A large variety in Dresses also contributes to the sparse sales. On an average, 20-30\% new attribute values are introduced for articles in a season. Because of these reasons we have decided not to forecast sales at variant-day level or article-day level. We are forecasting regular sales units at article-week level. The average number of regular sales units per article per store for Dresses in a season is 26 units and for Kids wear is 20 units.

Also, we saw for Dresses that in a season 50\% of the articles sell only for 4-5 weeks, 30\% of the articles sell for about 8-10 weeks and remaining 20\% sell for more than 10 weeks after the launch of the article. Similarly, for Kidswear 10\% of the articles sell for about 4–5 weeks, 40\% of the articles sell for about 8–10 weeks, 40\% of the articles sell for about 12-15 weeks and remaining 10\% articles sell for more than 15 weeks post launch. These statistics shows that Age, i.e., number of weeks since its launch is an important factor that impacts the sales of an article. In addition, we observed that similar articles in terms of attribute values, have similar sales pattern with respect to their age, which are shown in Figures \ref{age_dress} and \ref{age_kids}. The statistics shows that age of an article is extremely important and it impacts the sales. In Figures \ref{AGE_FIG} and \ref{fig:Feature}, sales units are scaled between 0 to 1 and are represented by Normalized sales.

The attributes of an item for both Dresses and Kids wear were also analyzed. Both of them share many common attributes such as sleeve length, fit, pattern, item length, material, fastening type, neck shape, color and item description. We found that color and sleeve length are some of the crucial attributes for demand forecasting, as can be seen from Figures \ref{color_dress}, \ref{sleeve_dress}, \ref{color_kids} and \ref{sleeve_kids}. These figures depict the mean normalised sales versus attribute values. From these figures we can easily infer that different attribute values have different sales contribution. 

The average selling price at which people buy Dresses is 25\$ and is 36\$ for Kids wear. Selling price of an item is a salient estimator for demand forecasting. The sales pattern with respect to price-bins is shown in Figure \ref{price_dress} and \ref{price_kids}, where value zero in the price bin represents low regular price range and value 4 in the price bin represents high regular price range. From these two Figures \ref{price_dress} and \ref{price_kids}, it is clear that sales has an inverse relationship with selling price and is thus used as an independent variable in our demand forecasting function.

	\section{ Methodology}
Demand forecasting for fashion items is a particularly challenging task owing to highly volatile demand. Traditional time series approaches use only the temporal features and are insufficient for accurately forecasting demand of fashion items. Moreover, for new fashion items with no historical sales data, time series based approaches cannot be used. Some approaches forecast demand based on similar visual characteristics. Since the similar visual attributes alone do not imply similar sales, these methods are less accurate. In this paper, we address the problem of accurate demand forecasting with a novel age based approach. In the age based model for demand forecasting of fashion items, the demand of an article $i$ in store $s$ at time $t$, is formulated as:

\begin{equation}
d_{i,s}= f( A_{i},t,Age_{i,s}, P_{i,s,t}, nholidays_t, S.I. )
\end{equation}
where,
\begin{itemize}
	\item $A_{i}$ is the attribute value corresponding to the article $i$. For e.g., attribute = color, material, pattern, usage and so on. 
	\item $t$ represents the number of time steps since the beginning of the season. We have considered one week as a time step.
	\item $Age_{i,s}$ is the number of time steps since the launch of the article $i$ in store $s$.
	\item $P_{i,s,t}$ is the selling price of an article $i$ in store $s$ at time step $t$.
	\item $nholidays_t$  represents the number of holidays from time step $t-1$ to time step $t$.
	\item  $S.I.$ represents the seasonality indices at a sub-class level.\\
\end{itemize} 

By using this formulation, we are collating the age, attribute values, temporal features and selling price of an article as demand estimators. This is done because the visual characteristics, selling price and the temporal features alone are insufficient to accurately predict the future demand.

\begin{figure}[ht!]
	\centering
	\includegraphics[scale=0.5,width=\linewidth]{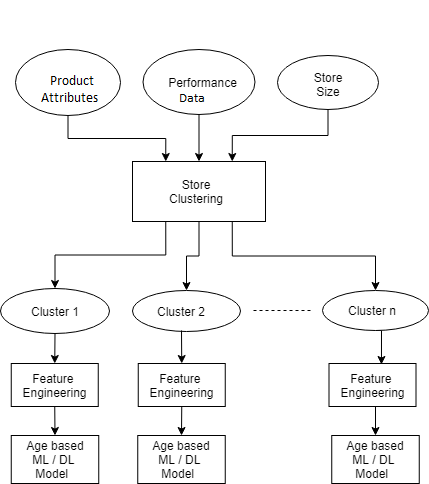}
	\caption{Training flow diagram for Age based model.}
	\label{Training_flow}
\end{figure}

\begin{figure}[ht!]
	\centering
	\includegraphics[scale=0.5,width=\linewidth]{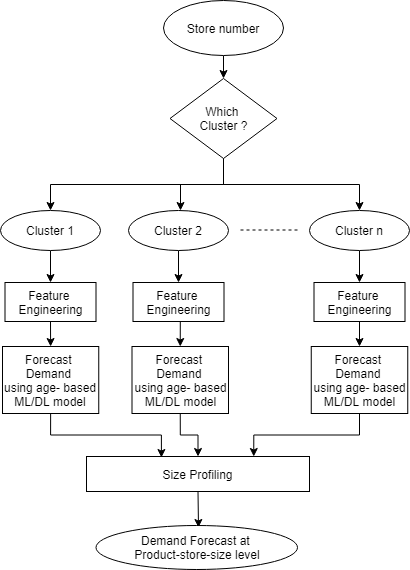}
	\caption{Testing flow diagram for Age based model.}
	\label{test_flow}
\end{figure}

	Figures \ref{Training_flow} and \ref{test_flow} present the training and testing framework of the age based forecast model along with the inputs and outputs.

With the available past performance and attribute data, we first perform store clustering. We propose an original method of clustering stores which reduces the intra-cluster variance of sales units. Conventionally, stores are grouped just by considering the size of the stores but similar sized stores do not imply similar sales behaviour. Our approach of clustering stores takes into consideration the sales units within each price-bin in a particular sub-class along with size of store. We observed that this approach of clustering captures the impact of demographic features such as mean income, age, and so on, even in the absence of such data. During the training phase, we obtain the store clusters. Our age based ML/DL models are trained within each cluster separately, as shown in Figure \ref{Training_flow}. At the end of the training phase, we obtain the trained age based ML/DL models for each cluster.

During the test phase, stores are mapped to different clusters based on their store identifiers as shown in Figure \ref{test_flow}. Within each cluster, the trained age based model is used to forecast demand at a article-store-week level. This is aggregated to obtain forecasts at phase and season level. After this step, the forecasted results are passed on to the size-profiler to obtain forecasts at variant level. 

\subsection{Feature Engineering}

\begin{figure}[ht!]
	\centering
	\includegraphics[width=\linewidth]{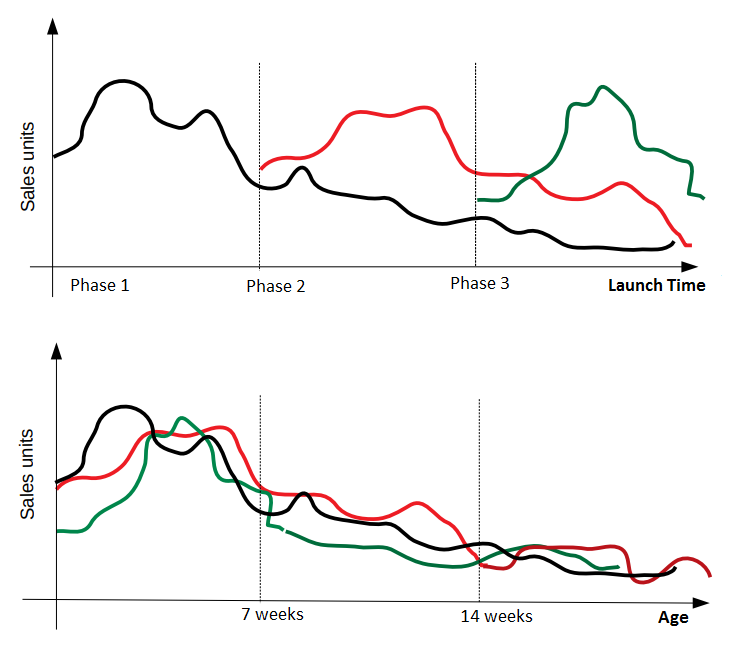}
	\caption{ Effect of translation from Launch time to Age.}
	\label{Age}
\end{figure}

\begin{itemize}
	\item \textbf{ Age }: The age of an article is calculated for each store, and is defined as the number of time steps from the time of its launch. We have considered one week as the time step. The newness of a fashion apparel is one of the significant contributors to its sales. We consider age as a monotonically increasing function of time. We use age along with other features, to give an estimate of the future demand. A commonplace practice for performing demand forecast for new items is to use attributes and corresponding sales data from the same time of the year in past year. In our approach, we left shift, in the time axis, the sales data s(t) to s(t+a), where a is the time of launch of the article after the season starts. Figure \ref{Age} shows the Age versus Sales plot. When this time shifted data is clubbed with the weekly seasonality indices we get a seasonal adjustment. This particular form of data perception helped us to train the model on a larger dataset, as now it considers the entire season data together and captures patterns from data across the season without risking the loss of seasonality factor.\\
	\item \textbf{ Number of holidays/special events}  : This feature captures the number of holidays and special events occuring within a time step. This feature captures holidays such as Easter. Valentines day and Mother's day are considered as special events and are included in the count of this feature. Including this feature in the model is important because our data analysis indicated a hike in sales around holidays and special events.\\
	
	\item \textbf{Color} : Attributes of an apparel include its color, material, pattern, sleeve length, neck shape, item length, fastening type and fit among others. A standard approach to use these categorical features for training a ML/DL model is to one-hot encode them. Another approach is to perform entity embedding on these categorical features. Our approach instead makes use of feature engineering of attributes in a way so as to bring out semantics from the categorical data available. The color attribute values have been transformed from categorical attribute values to a numeric vector. A vector for a particular color value comprises of the intensities of the red, green and blue component of the color. Encoding in this way helps in bringing the vectors representing colors with similar intensities closer to each other, while increasing the vector distance between colors of distant intensities. The lighter shades are now closer to each other and also farther from the darker shades than they would be in case of one-hot encoded vector.\\

	\item \textbf{Sleeve length} : This feature is specific to a Class and is included only in the Class possessing this attribute. We convert the sleeve length of an article to a numeric value based on the order of increasing sleeve length.\\
	
	\item \textbf{Usage} : Usage feature is derived from the item description for Women's Dresses, using Natural Language Processing technique. The usage of an article depicts the type of occasion the apparel is compatible with. We have considered the following usage types in Women's Dresses- casual, work, versatile, summer, tea, evening, beach, prom and party. When a dress is suited for more than one occasion, it comes under versatile.
\end{itemize}

This method is applied to structured data containing item attribute and performance characteristics. For implementing our age based demand forecasting approach, we use XGBoost and Deep learning models. Both these models are capable of capturing non-linear dependencies and scale well for large datasets and are thus suitable for our use case.

\subsection{Size Profiling }
Size profiling helps in estimating the quantity of variants in order to meet the store level demand. Proper size profiling leads to better customer satisfaction and retailers' profitablity. For size profiling, sales data of the previous year is first grouped at store-size level. We then use the ratio of sales contribution by each size within a store to cluster stores with similar size profiles. The mean cluster value is the final size profile for all the stores belonging to the cluster. This is based on the underlying assumption that the customer size profile in a region is consistent year on year. 

\subsection{Optimal time for launch of the article}
Retailers' revenue is influenced not only by demand forecasts but also by the item launch time. Our age based forecast model is also used for recommending the best time of launch for an article in the upcoming season. Articles can be launched at the start of each phase. Four different launch times are considered. We iteratively forecast demand for an article considering different launch times. We then select the best launch time for an article based on the criteria of maximizing revenue.

\subsection{Fashion Assortment}
In our use case, articles are launched at the begining of each phase within a season and so assortment optimization is run at the begining of each phase. The input to the assortment model is the forecasted sales, item attributes,  selling price and cost price of the article at a store level. The objective is to obtain the list and quantity of articles that is to be kept in the store for the particular phase, such that overall revenue is maximized and the business constraints are satisfied. The constraints are as follows:

\begin{itemize}
	\item The maximum distinct articles in a store.
	\item  Total budget at sub-class level.
	\item  Minimum and maximum distinct articles at sub-class level.
\end{itemize}

The assortment optimization is formulated as integer linear programming as follows:
\begin{equation}
\begin{matrix}
\textrm{Objective:}
\\\displaystyle \max & \multicolumn{3}{l}{\pi(x_{ij})=\displaystyle\sum_{i=1}^{|N|} \sum_{j=1}^{|M|} x_{ij}p_{ij}d_{ij}} \\
\end{matrix}
\end{equation}
\begin{equation}
\begin{matrix}
\textrm{subject\ to:}\\
\\
1.\ \ \ \ \ \ \displaystyle\sum_{i=1}^{|N|} x_{ij}& \leq & Count_j&\forall \ j \in M.\\
\\

2.\ \ \ \ \ \ \displaystyle\sum_{j=1}^{|M|}\sum_{i=1}^{|N|} x_{ij}d_{ij}c_{ij}I_k(i)&\leq & B_k &\forall \ k\in L.\\

3.\ \ \ \ \ \ \displaystyle\sum_{i=1}^{|N|} x_{ij}& \leq & Max\_Count_k&\forall \ k \in L.\\

4.\ \ \ \ \ \ \displaystyle\sum_{i=1}^{|N|} x_{ij}& \geq &Min\_ Count_k&\forall \ k\in L.\\
\\
5. \ \ \ \ \ x_{ij}&\geq&0 \ \ \ \ \ \forall \   i \in N\ and\ \forall j \in M. \ 
\end{matrix}
\label{opt}
\end{equation}

Where, 
\begin{itemize}
	\item $\pi$ is the total revenue.
	
	\item $M$ is set of stores.
	
	\item $N$ is set of articles.
	
	\item $L$ is set of sub-classes.
	
	\item$x_{ij}$ is a binary decision variable corresponding to an article i and store j.

	\item $d_{ij}$ is the predicted demand for an article i in store j.

	\item $p_{ij}$ is the selling price of an article i in store j.

	\item $c_{ij}$ is the cost price for an article i in store j.

	\item $Count_j$ is the maximum number of distinct articles that can be listed in store $j$.

	\item $Max\_Count_k$  is the maximum number of distinct articles for sub-class $k$.

	\item $Min\_Count_k$  is the minimum number of distinct articles for sub-class $k$.

	\item $B_k$ is the maximum budget for sub-class $k$.

	\item $I_k(i)$ is a binary indicator variable corresponding to sub-class $k$.

	\begin{equation}
	I_k(i)=\begin{cases}
	1, & \text{if $i\in  k$}.\\
	0,& \text{otherwise}.
	\end{cases}
	\end{equation}
	
\end{itemize}

\section{Experiments and Results}

\begin{figure*}[!ht] 
	\subfloat[Dresses]{%
		\includegraphics[width=1\textwidth]{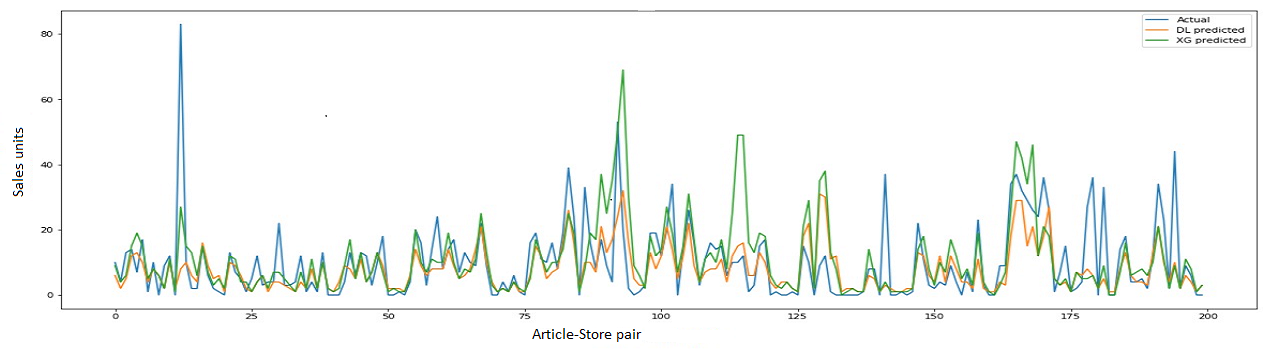} 
		\label{Dress_result}
	} 
	\\
	\subfloat[Kids wear ]{%
		\includegraphics[width=1\textwidth]{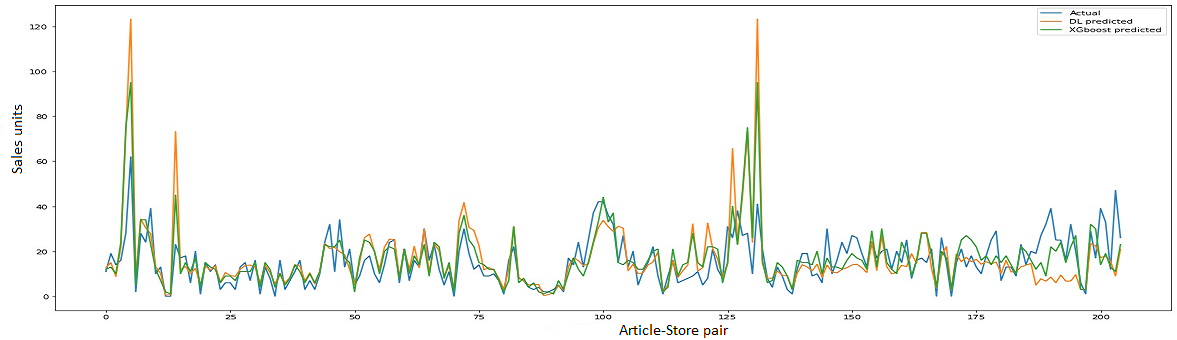}
		\label{kids_result}
	} 
	\caption{Forecast results for age based ML/DL models.}
	\label{fig:result}
\end{figure*}

	\begin{figure}[ht!]
	\centering
	\includegraphics[scale=0.8,width=\linewidth]{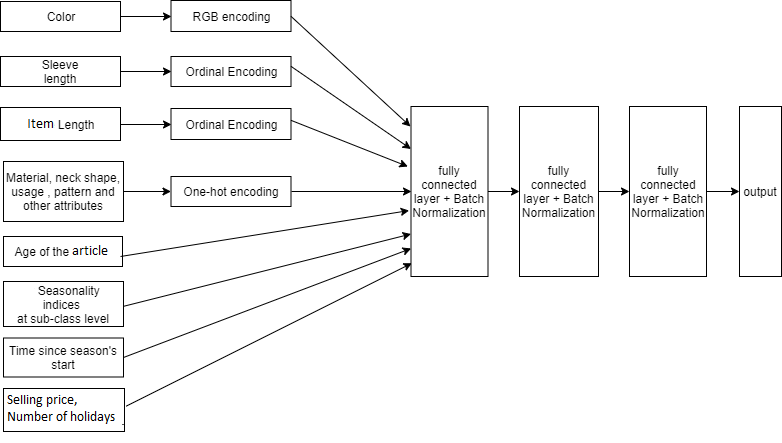}
	\caption{Architecture of the age based ANN model.}
	\label{ARCH}
\end{figure}

We demonstrate the robust performance of the approach through real world use case of a multinational fashion retailer. We have used historical data of the fashion retailer to train our models. The data constitutes performance, attributes and store characteristics of more than 300 brick and mortar stores. We have demonstrated results for- Dresses and Kids wear. We have chosen Dresses and Kids wear for demonstrations owing to their high contribution to sales. All the articles present during the test phase are new articles with no historical sales data. We have considered the spring-summer collection. The spring-summer season begins on the 1st of January and lasts till the end of June. We train our models using 6 months data from the same season in the previous year, 2018. After training, we forecast demand for the same season of next year, 2019, given the attributes data for the new articles. These forecasts are obtained at article-store-week level. Size profiling is performed to obtain forecast at size level. 

The age based ML/DL models for Dresses considers age of article, selling price, number of holidays/special events, seasonality indices, time since season starts, RGB encoded color values, Ordinal encoded sleeve length, Ordinal encoded item length and one-hot encoding of other attributes such as material, neck shape, usage, pattern, fit and fastening type as input. For Kids wear we consider the same input set except usage. 

We have used three different loss functions to train our age based models- 1) Mean Squared Error (MSE) 2) Logcosh 3) Huber loss. All these losses are concerned with the average magnitude of error regardless of the direction. MSE heavily penalizes larger errors in comparison to less deviated predictions. Huber loss is more robust to outliers as compared to MSE. Logcosh works almost similar to the MSE except that it is less severely affected by the occasional extreme errors.
Also, to comapre the performance of the different demand forecasting models, we have used Root Mean Squared Error (RMSE) and Mean Absolute Error (MAE) as evaluation metrics.

Deep learning models are built using tensorflow deep learning framework \cite{tensorflow2015-whitepaper}. XGBoost models have been built using the scikit-learn package \cite{scikit-learn}. Bayesian-Hyperparameter Optimization technique \cite{bergstra2013making} is used for hyperparameter tuning of XGBoost. We have implemented  ANN with three dense layers along with batch-normalization \cite{santurkar2018does} and dropout \cite{srivastava2014dropout}. Architecture of the age based ANN model is shown in Figure \ref{ARCH}. We have trained our ANN using the Adam optimizer with a learning rate of 0.001. We have scaled our approach to run for each category of the fashion retailer in a distributed computing environment.

We have used the average model as a baseline for comparison of our results. In the average model, we first identified similar articles from the previous year during the same season. We have considered selling price and ranked item attributes to obtain top 3 similar articles using nearest neighbour algorithm. Attribute's significance  was obtained as an input from the business and higher weight was given to higher ranked attributes. After obtaining the set of similar articles, we have taken the average sales units of all the top three similar articles in the given time step and used that as an estimate for the demand of the corresponding new article. The results obtained from our baseline average model are tabulated in Tables \ref{Dresses_table} and \ref{kids_table}. As we can observe from these tables, this approach did not yield satisfactory results. The high error implies that the sales of similar articles were not close enough to sales of the new article. This result can be attributed to the fact that the similar articles for a given article might have been launched across different phases in the previous year's season resulting in differences in seasonality factor.

\begin{table*}[ht!]
	\caption{Dresses Results}
	\centering
	\small\addtolength{\tabcolsep}{-5.4pt}
	\begin{tabular}{|l|l|l|l|l|l|l|l|l|}
		\hline
		\multicolumn{2}{|l|}{\multirow{2}{*}{}}                              & \multirow{2}{*}{\textbf{Loss Function}} & \multicolumn{3}{l|}{\textbf{RMSE}}                     & \multicolumn{3}{l|}{\textbf{MAE}}                      \\ \cline{4-9} 
		\multicolumn{2}{|l|}{}                                               &                                         & Article-Store-Week & Article-Store-Phase & Article-Store-Season & Article-Store-Week & Article-Store-Phase &Article-Store-Season \\ \hline
		\multicolumn{3}{|l|}{\textbf{Average model}}                                                                   & 6.83             & 13.62            & 39.20              & 3.54            & 8.09             & 26.80              \\ \hline
		\multicolumn{2}{|l|}{\textbf{Baseline Xgboost}}                     & MSE                                     & 5.14                &12.36                  &34.70                   &3.18                 &7.82                  &22.93                   \\ \hline
		\multicolumn{2}{|l|}{\textbf{Baseline DL}}                          & MSE                                     &5.21                 &11.93                  &32.12                   &2.99                &7.53                  &21.38                  \\ \hline
		\multirow{6}{*}{\textbf{Age Based Model}} & \multirow{3}{*}{Xgboost} & MSE                                     & 1.98            & 9.89             & 22.34             & 1.61            & 5.71             & 12.13             \\ \cline{3-9} 
		&                          & HUBER                                   & 4.23            & 10.36             & 23.71             & 2.48            & 7.95             & 12.13             \\ \cline{3-9} 
		&                          & logcosh                                 & 3.95            & 8.29             & 21.17             & 2.37            & 7.63             & 11.46             \\ \cline{2-9} 
		& \multirow{3}{*}{DL}      & MSE                                     & 2.86            & 9.67             & 22.79             & 2.71            & 6.84             & 15.67             \\ \cline{3-9} 
		&                          & HUBER                                   & 4.09            & 10.81             & 25.12              & 2.64            & 7.93             & 15.43             \\ \cline{3-9} 
		&                          & logcosh                                 & 4.21             & 9.55             & 24.93              & 2.64            & 6.05             & 15.78             \\ \hline
	\end{tabular}
	\label{Dresses_table}
\end{table*}

\begin{table*}[ht!]
	\caption{Kids wear Results}
	\centering
	\small\addtolength{\tabcolsep}{-5.4pt}
	\begin{tabular}{|l|l|l|l|l|l|l|l|l|}
		\hline
		\multicolumn{2}{|l|}{\multirow{2}{*}{}}                              & \multirow{2}{*}{\textbf{Loss Function}} & \multicolumn{3}{l|}{\textbf{RMSE}}                     & \multicolumn{3}{l|}{\textbf{MAE}}                      \\ \cline{4-9} 
		\multicolumn{2}{|l|}{}                                               &                                         & Article-Store-Week & Article-Store-Phase & Article-Store-Season & Article-Store-Week & Article-Store-Phase & Article-Store-Season \\ \hline
		\multicolumn{3}{|l|}{\textbf{Average model}}                                                                   & 3.56            & 10.55            &39.27             &3.07            & 7.25             &28.8              \\ \hline
		\multicolumn{2}{|l|}{\textbf{Baseline Xgboost}}                              & MSE                                     & 2.97                &7.61                  &31.89                   & 2.53                &6.13                  &22.76                   \\ \hline
		\multicolumn{2}{|l|}{\textbf{Baseline DL}}                                   & MSE                                     & 2.84                & 7.63                 &30.23                   &2.61                 &6.11                  & 21.31                  \\ \hline
		\multirow{6}{*}{\textbf{Age Based Model}} & \multirow{3}{*}{Xgboost} & MSE                                     & 1.65            & 4.92             & 15.38             & 1.21             & 2.83              & 7.12              \\ \cline{3-9} 
		&                          & HUBER                                   & 1.64            & 4.96             & 15.43             & 1.19            & 2.68             & 6.93              \\ \cline{3-9} 
		&                          & logcosh                                 & 1.61            & 4.89             & 15.41              & 1.17            & 2.54             & 6.85              \\ \cline{2-9} 
		& \multirow{3}{*}{DL}      & MSE                                     & 1.54            & 4.32             & 13.04             & 1.15            & 2.41             & 6.53              \\ \cline{3-9} 
		&                          & HUBER                                   & 1.56            & 4.35             & 13.15             & 1.21            & 2.34             & 6.68              \\ \cline{3-9} 
		&                          & logcosh                                 & 1.51            & 4.29             & 12.95             & 1.13            & 2.35             & 7.34             \\ \hline
	\end{tabular}
	\label{kids_table}
\end{table*}

\begin{table*}[ht!]
	\centering
	\small\addtolength{\tabcolsep}{1pt}
	\caption{Ideal Launch Time}
	\begin{tabular}{|l|l|l|l|l|l|}
		\hline
		\textbf{Store id} & \textbf{Article id} &\textbf{Demand for phase 1} &\textbf{Demand for phase 2} &\textbf{Demand for phase 3} & \textbf{Ideal launch phase}  \\ \hline
		1111     & 206021      & 25                & 19                & 17                & 1                  \\ \hline
		1111     & 206049      & 33                & 25                & 13                & 1                  \\ \hline
		1122     & 206049      & 26                & 21                & 19                & 1                  \\ \hline
		1122     & 206045      & 10                & 8                 & 6                 & 1                  \\ \hline
		1111     & 601755      & 60                & 38                & 40                & 1                  \\ \hline
		1111     & 601763      & 28                & 18                & 15                & 1                  \\ \hline
		1133     & 224806      & 91                & 104               & 83                & 2                  \\ \hline
		1132     & 601238      & 30                & 40                & 25                & 2                  \\ \hline
		1111     & 601291      & 40                & 100               & 80                & 2                  \\ \hline
		1155     & 601326      & 72                & 99                & 86                & 2                  \\ \hline
		1188     & 600913      & 26                & 39                & 19                & 2                  \\ \hline
		1122     & 601236      & 21                & 35                & 24                & 2                  \\ \hline
		1100     & 601238      & 25                & 38                & 31                & 2                  \\ \hline
		1122     & 224811      & 13                & 18                & 11                & 2                  \\ \hline
		1144     & 224666      & 24                & 39                & 45                & 3                  \\ \hline
		1144     & 601291      & 21                & 18                & 37                & 3                  \\ \hline
		1144     & 601292      & 6                 & 9                 & 10                & 3                  \\ \hline
		1166     & 224667      & 3                 & 7                 & 8                 & 3                  \\ \hline
		1166     & 224668      & 3                 & 7                 & 8                 & 3                  \\ \hline
	\end{tabular}
	\label{Launch}
\end{table*}

In Tables \ref{Dresses_table} and \ref{kids_table},  we have also summarized the results of the baseline Xgboost and ANN models. These baseline models consider the item attributes, which are one-hot encoded, as a feature along with selling price and time since the beginning of season. The derived feature considered in the baseline models is number of holidays. Also, in the baseline model, we perform one-hot encoding of the color feature. Note that we do not consider age, RGB encoded color and usage as a feature in the baseline models. From the results presented in Tables \ref{Dresses_table} and \ref{kids_table}, it is evident that age based forecast models are more accurate than the  baseline Xgboost and ANN models.

We observe from Tables \ref{Dresses_table} and \ref{kids_table} that  both our age based ML and DL models outperform the average prediction model, baseline Xgboost and the baseline DL model. This can be attributed to the fact that similar items across the season are captured along with their newness and the sub-class seasonality index in the age based forecast models. 

From Figure \ref{kids_result}, we observe the actual versus predicted sales units of Kids wear for randomly selected article-store pair in a phase. We observe few instances of under and over predictions, but for most of the store-article combination, we obtain good demand forecasts. The demand forecast results for Dresses for randomly selected article-store pair in a phase are shown in Figure \ref{Dress_result}. We observe that the age based ML/DL models perform remarkably well in both Dresses and Kids wear. Demand forecast for Women's Dresses is comparatively more challenging, which is also intuitively justified, and can be observed from Figures \ref{Dress_result} and \ref{kids_result}.

\begin{figure}[ht!]
	\centering
	\includegraphics[width=\linewidth]{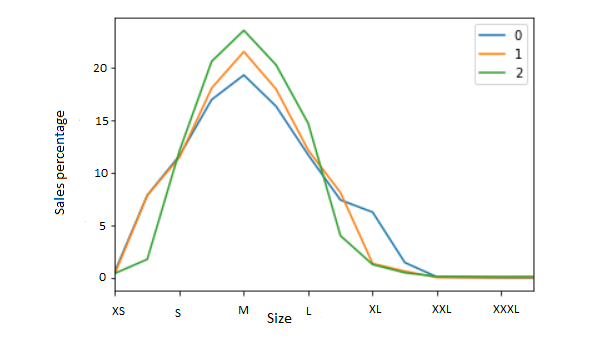}
	\caption{ Size Profile for Dresses.}
	\label{size_profile}
\end{figure}

We have obtained three different size profiles for Dresses as shown in Figure \ref{size_profile}. In Figure \ref{size_profile}, the different plot legends labelled 0, 1 and 2 represents three different size profiles across the stores. We observe that although the three size profiles have almost similar distribution, there is some variation in percentage share for each size. Similarly, we have obtained three different size profiles for Kids wear.

Table \ref{Launch} demonstrates an anonymised sample data from the fashion retailer. For each article-store combination, we obtain the number of units forecasted if launched in phase 1, phase 2, and phase 3. We then select the best launch phase based on the criteria of maximizing revenue. From Table \ref{Launch}, we observe that launching articles at different phases results in different sales units by the end of the season.

We used our age based demand forecasts to obtain the store level assortment for the next season in 2019. We calculated the total revenue corresponding to our assortment. Similarly, we obtained the retailer's total revenue based on his forecasts and his assortment plan for the season in 2019. We then compared our model's revenue values with the retailer's revenue numbers from his plan. We found a revenue upliftment of 41\% from our model as compared to the retailer's plan.

\section{Conclusion and Future Work}
In the fashion industry, a forecast is not a stand-alone entity and retailers have to exploit the potential insights offered by it to drive intelligent decisions in their supply chain. For example, the fashion retailers' problem is seeing how these forecasts can be leveraged in supply chain for ordering from vendors, assortment planning, inventory planning and determining new item launch period, so as to maximize their total revenue.

In this paper, we have provided solutions for addressing all these problems and not only the stand-alone solutions. We have shown the implementation of a novel age based prediction model to accurately forecast demand for fashion items for the next season. We have also introduced novel feature engineering for fashion forecast by using RGB component in place of color category. We demonstrate the robust performance of the approach through real world use case of a multinational fashion retailer. We also show that our prediction model can be leveraged to recommend the ideal time of launch for each new item in the next season. This cold start problem is a very important issue that crops up for the retailer every year. Our prediction model provides insights and actionable strategies for the retailer for planning the launch dates for the new items for the next season. Our fashion assortment optimization model produces list of items and number of units to be listed in a store for next season that maximizes total revenue, given demand forecasts and business constraints as inputs. When these forecasts were taken in tandem with the assortment planning model we developed, we found that our solution showed an average uplift of 41\% in revenue when compared to the retailer's plan. Our solution leads to better inventory planning, assortment planning, ordering and overall increase in profit for the fashion retailer's supply chain. Our model is being extended and deployed to all other categories for the retailer.

Other than age, we found that other hidden variables impact sales, such as mothers mostly buying for their families. We intend to identify and feature engineer a mixture of these variables and forecast. We also plan to include data for competitor, customer, social media, weather, number of changing rooms in a store, number of footfalls and qualitative factors such as future fashion trends. We will be including data from ecommerce, promotion, market and store attributes. We also intend to use images of the items and use image related features in our model.

\bibliographystyle{ACM-Reference-Format}
\bibliography{sample-base}

\end{document}